\documentclass{article}

\usepackage[final]{nips_2018}

\usepackage[utf8]{inputenc} 
\usepackage[T1]{fontenc}    
\usepackage{hyperref}       
\usepackage{url}            
\usepackage{booktabs}       
\usepackage{amsfonts}       
\usepackage{nicefrac}       
\usepackage{microtype}      
\usepackage{graphicx}

\title{Time Aggregation and Model Interpretation for Deep Multivariate Longitudinal Patient Outcome Forecasting Systems in Chronic Ambulatory Care}

\author{
  Beau Norgeot\textsuperscript{1}  \and Dmytro Lituiev\textsuperscript{1}  \and Benjamin S. Glicksberg\textsuperscript{1} \and Atul J. Butte\textsuperscript{1*} \and \\
  \textsuperscript{1} Bakar Computational Health Sciences Institute, University of California, San Francisco \\
  \texttt{*atul.butte@ucsf.edu}
}

\begin{document}

\maketitle

\begin{abstract}
Clinical data for ambulatory care, which accounts for 90\% of the nations healthcare spending, is characterized by relatively small sample sizes of longitudinal data, unequal spacing between visits for each patient, with unequal numbers of data points collected across patients. While deep learning has become state-of-the-art for sequence modeling, it is unknown which methods of time aggregation may be best suited for these challenging temporal use cases. Additionally, deep models are often considered uninterpretable by physicians which may prevent the clinical adoption, even of well performing models. We show that time-distributed-dense layers combined with GRUs produce the most generalizable models. Furthermore, we provide a framework for the clinical interpretation of the models.
\end{abstract}

\section{Introduction}
Chronic ambulatory disease care is associated with the overwhelming majority of death, disability, and healthcare spending \cite{buttorff2017multiple}, \cite{centers2016national} in the United States. Successful predictive modeling in this setting has the potential to significantly improve clinical care, patient quality of life, and healthcare efficiency. Identifying the best methods and functions for time aggregation will be a critically important part of building models that perform well in these types of settings. However, even well performant models may not be sufficient to warrant the clinical adoption of artificial intelligence (AI) to longitudinal patient care. Deep learning has traditionally been met with resistance in the clinical community due to a general sentiment that the models function entirely as uninterpretable black boxes. While we agree that the use of black boxes in clinical care should be avoided wherever possible, we posit that deep time series models need not be black boxes at all. By transferring and extending a traditional method to calculate variable importance in machine learning models known as Permutation Importance Scoring \cite{breiman2001random}  to deep time series modeling and by developing a method of visualizing the final model-learned patient representations as clusters we aim to show that it is possible to interpret the driving factors behind model predictions on both the patient and population levels. These methods can not only contribute to model interpretation but could be used in the future to generate and test medical and pharmaceutical hypotheses in this space in which individual progression and response to treatment for many of the diseases may not be well understood. 

We selected Rheumatoid Arthritis (RA) as a use case. RA is a common (~1\% nationwide) complex chronic autoimmune disease with unknown causes along with highly individualized responses to therapeutics and disease progression. It is associated with significant morbidity and a high cost of care. Our goal was to examine the impact of time aggregation strategies on deep time series models which used a patient’s history of labs, medications, and disease activity along with their current treatment plan and current clinical measurements to forecast whether a patient’s disease activity would be controlled or uncontrolled at their next visit. The highest performing model was then examined for interpretation using the approaches described above.

\section{Methods}
\label{gen_inst}

Electronic Health Record (EHR) data were extracted from two rheumatology clinics with significantly different patient populations and provider treatment patterns, a University Clinic (UC) and a public Safety Net (SN). Patients from the larger UC cohort (n=578) were split into three groups [train (n=369), validation (n=93), test (n=116)] using stratified random sampling on the primary outcome which was the binary category of controlled or uncontrolled disease activity at their most recent clinical visit. Patients from the smaller SN cohort (n=242) were split into two groups, train (n=125) and test (n=117) using stratified random sampling as previously described. The patient data were grouped into three windows of one hundred days each (which corresponds to  the the median number of days between visits) to overcome unequal length of time between visits. Data within each window was further aggregated, selecting the most recent value of any variable with more than one value within a given window to adjust for unequal numbers of data points between patients within and across widows.  

\subsection{Time Aggregation Functions}
To better understand the impact of various time aggregation functions on model performance we designed Bayesian Optimization experiments (via Hyperas a Hyperopt \cite{bergstra2013hyperopt} wrapper for Keras \cite{chollet2015keras} which itself wraps Tensorflow \cite{abadi2016tensorflow} ) to examine the efficacy of six different time aggregation function approaches. The time aggregation approaches that we selected were: fully dense, time distributed dense (TDD) followed by a dense layer, TDD followed by valid-padded convolution, TDD followed by causal convolution, TDD followed by LSTM, TDD followed by GRU. 

For simplicity, we constrained each architecture to consist of one input layer (which could itself function as a time aggregation layer), one time aggregation layer, and a dense layer prior to the output. Models were checkpointed by monitoring validation loss to control for variations associated with training time for each layer type. All models used Adam (with the default parameters) as the optimization function and a batchsize of  64 samples. The Bayesian Optimization selected hyperparameters for each time aggregation function network by minimizing binary cross entropy loss including: number of units and L1/L2 regularization at each layer, and proportion of drop out between layers.  We then generated confidence intervals for the best performing model for each time aggregation function using the Delong Method \cite{delong1988comparing}. Models were trained on the UC train cohort and tested against the UC validation cohort. 
We selected the top performing time aggregation architecture and then trained on the combination of UC train and validation cohorts. This became the fully trained model that was used for the Interoperability methods; Permutation Importance Scoring and Confusion Plots.

\subsection{Interpretability}
\subsubsection{Longitudinal Permutation Importance Scores across Windows}
Using data from the UC cohort, we calculated longitudinal permutation importance score for all input variables (medications, quantitative inflammatory markers, and clinical scores) across all time points within the test set with respect to auROC, to estimate the effect of each variable at each time point on predictive accuracy . We began with the original inputs for the test set. For each variable in each time step in the test set, we replaced the given variable within the given time window with a random value sampled from the same variable and the same time window from training set. Using the original inputs with just a single randomly replaced variable value for each patient, we used the fully trained model to re-calculate predicted disease probabilities and then generated an auROC based on those probabilities. We repeated the random sampling 20 times for each variable-window combination, and recorded the mean auROC across the sampling rounds. Finally, we visualized the relative difference between the mean auROC obtained in permutation from the auROC on original test set as a heat map graph.

Relative Difference = (mean(permuted score) - original score) / (original score)

\subsubsection{Population Differences}
We generated a graphic, which we have called a Confusion Plot, by extracting the final dense representation learned by the fully trained model for each patient and plotting them using T-SNE, colored by outcome category, to assess the coherence of the representations learned by the model. We performed this experiment for each cohort and compared the results side by side to determine the differences in patient representation for the model in each patient population. 

\section{Results}

\subsection{Time Aggregation}
Results from the time aggregation experiment can be found in Table~\ref{Table-1}. Using a Time Distributed Dense (TDD) layer as the input layer provides the single largest increase in predictive performance as seen by the relative difference between the TDD and Dense architectures. Using a convolutional layer with non-causal padding after the TDD provides a modest improvement over a dense layer, while using causal padding increases performance even further. The top performing architectures used a recurrent layer following the TDD with GRUs outperforming LSTMs. 

\begin{table}[h]
	\scriptsize
	\caption{Time Aggregation Experiment Results}
	\label{Table-1}
	\centering
	\begin{tabular}{lllllll}
		\toprule
		\cmidrule(r){1-2}
		Function&Dense&TDD Dense&TDD GRU&TDD LSTM&TDD CNN&TDD Causal CNN \\
		\midrule
  	AUC & 0.778 & 0.817 & 0.845  & 0.838 & 0.821  & 0.832  \\
  	95\% CI & [0.683, 0.864]  & [0.731, 0.894] &  [0.753 ,0.914] &[0.743 ,0.911] & [0.727, 0.897] &  [0.740, 0.906] \\
		\bottomrule
	\end{tabular}
\end{table}

\subsection{Interpretability}
\subsubsection{Longitudinal Permutation Importance Scoring}
More recent time windows are more important than more distant time windows. CDAI history is of greatest importance followed by steroid prescription (Prednisone, a reasonable surrogate for all steroids representing 55\% of steroids prescribed in our data set). Changing a patient's previous DMARD treatment strategy at the visit prior to the most recent visit was of significant performance as were the presence or absence of certain specific DMARDs.

\begin{figure}[h]
	\centering
	\includegraphics[scale=0.5]{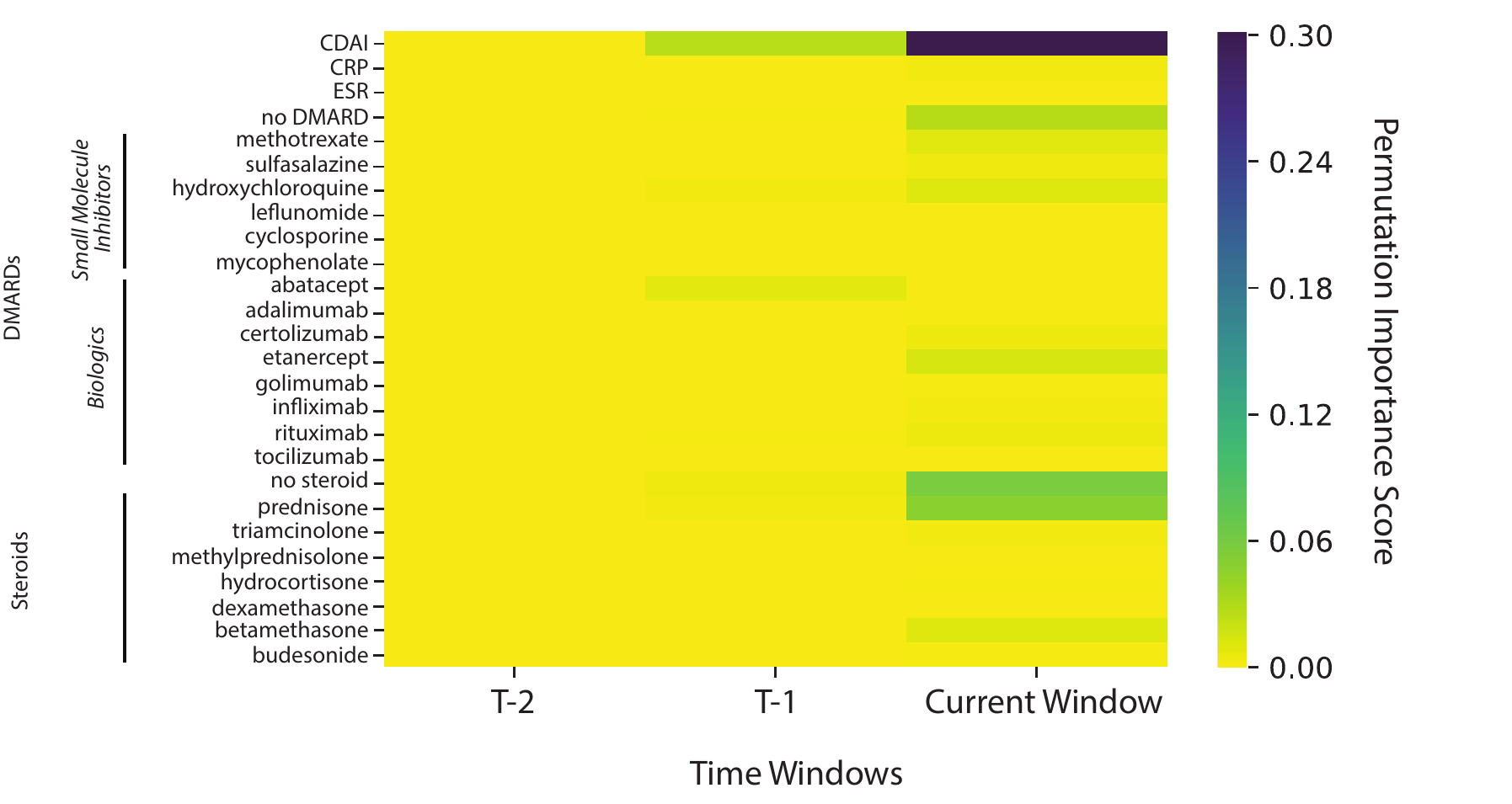}
	\caption{Permutation Importance Scores}
	
\end{figure}

\begin{figure}[h]
	\centering
	\includegraphics[scale=0.5]{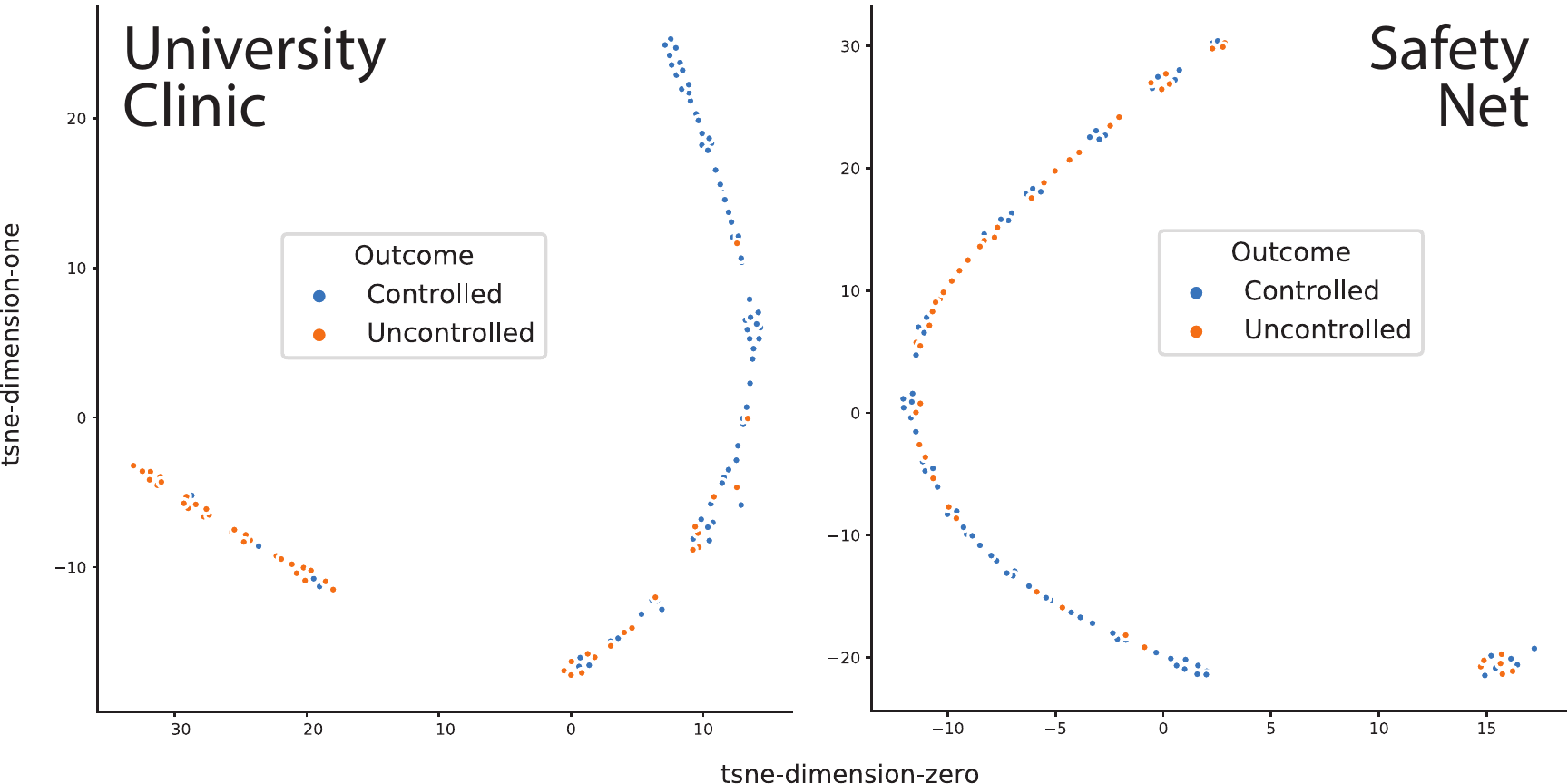}
	\caption{Confusion Plots}
	
\end{figure}

\subsubsection{Confusion Plot }
In both the UC and SN populations the final patient representations that the model learned formed a one-dimensional manifold (a curve). The model is clear in both clinics about patients that will definitely be at one end of the disease activity spectrum or the other at their next clinical visit. However, the confusion plots are clearly different between the two populations. In the UC cohort, the presence of Controlled and Uncontrolled patient representations mix in just one pocket of close proximity to each other in the middle of manifold, while within the SN Clinic cohort there are multiple of these highly proximus representations that begin to occur in pockets much closer to the tails.

\section{Discussion and Conclusions}
\label{others}

In this study, we compared different time aggregation functions for ambulatory outcome forecasting and provided a framework for interpreting models in this setting. We found that using a time distributed dense layer (which uses the same function to re-weight input features across all time windows), followed by recurrent modeling of the re-weighted windows produced the best results. 

Longitudinal Permutation Importance Scoring reveals that newer time points are most important, and that recent disease activity scores are important but that quantitative inflammatory markers are not. Prescription of new steroids at the current visit, which we interpret to clinically indicate currently uncontrolled disease, and maintaining or switching to a new DMARD, we interpret to act as surrogate for the patient and physician believing that the current DMARD is working, are also important.

There are some findings such as the influence of certain DMARD changes and steroids additions that require further examination to determine whether these finding may lead to new treatment strategies or are the result of confounding by indication. A potential limitation of the Longitudinal Permutation Importance Scores as implemented in this paper is a lack of directional effect between future disease activity and each variable within each window. This could be solved be reporting the changes in probability instead of the changes in auROC associated with each permutation, though additional considerations will need to be made for the continuous variables. An additional straightforward application of directional longitudinal permutation importance scoring would be to permute medication choices to optimize probabilities for a successful outcome. 

Examining the Confusion Plots in the current study, the single mixed patient pocket at the UC seemed to indicate a natural transition between patients who have Controlled and Uncontrolled disease state at their next visit, while the multiple mixed pockets for the SN patients perhaps indicates that strong confounding factors are driving outcomes for many patients. Since the patients in the SN cohort, unlike those in the UC, are known to be predominantly non-White and be considerably less likely to have private insurance, these findings support suspected social determinants of health within the population that should be further examined.

In summary, Longitudinal Permutation Importance Scores can be extended from traditional machine learning approaches into longitudinal deep learning methodologies, providing insight into variable significance over time. Confusion Plots, which visualize model-learned dense patient representation vectors, can used to search for sub-cohorts, indicate the presence of potential confounding factors, and examine differences between subgroups and populations. Taken together, we found that longitudinal deep learning can be successfully applied to ambulatory disease forecasting and that the resulting models can be interpreted in a straightforward manner. We expect that these methods will be used to facilitate the adoption of deep learning in the field of clinical medicine.

\bibliography{nips_2018}

\end{document}